# Exploring Hierarchical Classification Performance for Time Series Data: Dissimilarity Measures and Classifier Comparisons

Celal ALAGOZ

Kırıkkale, Türkiye.

https://orcid.org/0000-0001-9812-1473

celal.alagoz@gmail.com

**ABSTRACT**

The comparative performance of hierarchical classification (HC) and flat classification (FC) methodologies in the realm of time series data analysis is investigated in this study. Dissimilarity measures, including Jensen-Shannon Distance (JSD), Task Similarity Distance (TSD), and Classifier Based Distance (CBD), are leveraged alongside various classifiers such as MINIROCKET, STSF, and SVM. A subset of datasets from the UCR archive, focusing on multi-class cases comprising more than two classes, is employed for analysis. A significant trend is observed wherein HC demonstrates significant superiority over FC when paired with MINIROCKET utilizing TSD, diverging from conventional understandings. Conversely, FC exhibits consistent dominance across all configurations when employing alternative classifiers such as STSF and SVM.

Moreover, TSD is found to consistently outperform both CBD and JSD across nearly all scenarios, except in instances involving the STSF classifier where CBD showcases superior performance. This discrepancy underscores the nuanced nature of dissimilarity measures and emphasizes the importance of their tailored selection based on the dataset and classifier employed.

Valuable insights into the dynamic interplay between classification methodologies and dissimilarity measures in the realm of time series data analysis are provided by these findings. By elucidating the performance variations across different configurations, a foundation is laid for refining classification methodologies and dissimilarity measures to optimize performance in diverse analytical scenarios. Furthermore, the need for continued research aimed at elucidating the underlying mechanisms driving classification performance in time series data analysis is underscored, with implications for enhancing predictive modeling and decision-making in various domains.

## INTRODUCTION

Many datasets exhibit a hierarchical structure where data or objects are organized into a tree-like arrangement with nested categories or groups. The adaptation of a classification scheme specific to such hierarchies is known as HC, demonstrating its effectiveness across various fields (Silla and Freitas, 2011). In situations where the hierarchy is not predefined, extracting it from multi-class datasets and transforming a FC task into HC has proven advantageous in terms of both efficiency and efficacy. This approach has found success in diverse domains, including image, text, and multivariate data analysis (Punera et al., 2006; Bengio et al., 2010; Helm et al., 2021; Punera et al., 2005; Punera et al., 2006). This technique, referred to as automated hierarchy generation or learned hierarchy, involves extracting hierarchy from flat labels.

However, despite the demonstrated effectiveness in various domains, the exploration of automated hierarchy generation for time series data, specifically in terms of learning hierarchy from time series data and transforming FC tasks into HC using the generated hierarchy, remains largely uncharted. Investigating this novel aspect contributes valuable insights to the field.

The foundational assumption behind HC is the existence of dependencies between classes in a multi-class dataset. In real-world scenarios, data points are associated with classes rather than the classes themselves, necessitating a representation of classes based on data points. This representation can be attained using semi-supervised methods, wherein intrinsic properties of data points are explored to infer class relations, leveraging known class labels of data points. Alternatively, supervised approaches employ extrinsic models to examine the distinguishability of classes or class groups and aid in extracting structural dependencies between them.

Unsupervised representation methods involve obtaining class-conditional centroids, such as class-conditional means, or using class-conditional distributions to represent classes with associated data points. Once representations are established, the next step involves determining a distance metric or, more broadly, a dissimilarity measure to quantify the structural relations between classes. Standard distance metrics can be applied when classes are represented as class-conditional centroids. For class-conditional distributions, information-theoretical measures are employed to assess dissimilarity between distributions. Notably, Punera et al. (2006) utilized JSD to automatically generate hierarchy, while Helm et al. (2021) employed Task Similarity Distance TSD.

Exploring the dependence between classes in a supervised manner entails running a classifier on the training data and evaluating its performance to estimate the distinguishability of classes. Class similarities are derived from the prediction performance of the classifier using a confusion matrix (Godbole, 2002) or an extended version that integrates class estimations and class probability estimations (Silva-Palacios et al., 2018).

This study aims to fill the gap in the literature by investigating the transformation of flat-labeled multi-class time series datasets into hierarchical forms. It does so by learning hierarchy with dissimilarity measures, specifically JSD, TSD, and CBD, which utilizes a confusion matrix with estimated classes to derive a separability measure between class pairs.

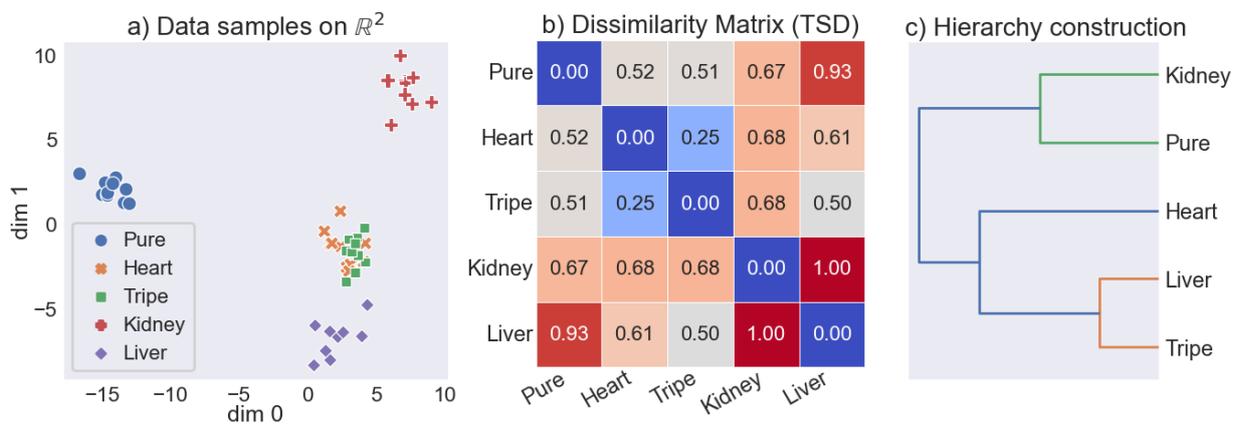

Figure 1. Hierarchy generation steps for the Beef dataset are demonstrated. For illustrative purposes, data instances are visualized in a 2D space, as shown in (a). Subsequently, a dissimilarity measure is obtained (b), and, finally, a hierarchy is built using a divisive clustering approach that employs dissimilarity measures as input to the clustering algorithm, K-medoids (c).

**MATERIALS AND METHODS**

The process of transforming a FC task into HC involves two key stages: hierarchy generation and hierarchy exploiting. Such transformation is applied to a dataset denoted as $D$, consisting of $c$ classes (or flat labels) and $N$ samples. Each sample represent univariate, fixed-length time series with $L$ time points, where $D = \{(x_i, y_i)\}_{i=1}^{N}$ and $x_i = \{x^j\}_{j=1}^{L}$ while $y_i \in \{1,2,\ldots,c\}$.

Hierarchy generation stages employed in this study are illustrated in Figure 1, using the "Beef" dataset as a case study. he hierarchy generation stages employed in this study are illustrated in Figure 1, using the "Beef" dataset as a case study. A dissimilarity measure - either JSD, TSD, or CBD - is obtained from the training set of data samples. To ascertain the structural dependence between classes, the associated labels of data points are needed. Pairwise dissimilarity measures between classes (cxc) are estimated, forming a dissimilarity matrix. Classes are then embedded into a vector space using multi-dimensional scaling (Wickelmaer, 2003) or spectral embedding (Von Luxburg, 2007). Alternatively, this study utilizes clustering directly from the pairwise dissimilarity matrix via K-Medoids, which offers a precomputed distances option. The hierarchy building stage is implemented once class separability is ensured through the dissimilarity measure. Hierarchy building can occur in two ways: divisive clustering or agglomerative clustering. This study adopts a divisive clustering scheme employing K-Medoids for the partitioning task.

Regarding hierarchy generation, the goal is to learn a hierarchy from available data with the highest possible quality. Although hierarchy quality is not a straightforward concept, exploiting the separability of classes is fundamental. The shared objective of dissimilarity measures is to place easily separable classes in the hierarchy as far from each other as possible, and vice versa. The following subsections introduce the dissimilarity measures JSD, TSD, and CBD, which are employed in this study.

**Jensen-Shannon Divergence:**

JSD is a measure of dissimilarity between two probability distributions (Lin, 1991). It is a symmetric and smoothed version of the Kullback–Leibler Divergence (KLD) (Kullback and Leibler, 1951), addressing some of KLD's limitations, such as sensitivity to sparse data and asymmetry. It is commonly used in information retrieval, machine learning, and bioinformatics for tasks such as document clustering, text classification, and comparing probability distributions

The JSD between two probability distributions, P and Q, is calculated as follows:

$$JSD(P,Q) = \frac{1}{2}(D_{KL}(P \parallel M) + D_{KL}(Q \parallel M)) \qquad [1]$$

where $D_{KL}(P \parallel M)$ denotes the Kullback–Leibler Divergence between distributions P and Q, and M is the average of P and Q:

$$M = \frac{1}{2}(P + Q) \qquad [2]$$

The JSD is bounded between 0 and 1, where 0 indicates identical distributions, and 1 signifies completely dissimilar distributions. JSD can be generalized to compare more than two classes/distributions such as:

$$JSD_{\pi_1,\ldots,\pi_c}(P_1, P_2, \ldots, P_n) = \sum_i \pi_i D_{KL}(P \parallel M) = H(M) - \sum_{i=1}^{n} \pi_i H(\pi_i) \qquad [3]$$

where $M := \sum_{i=1}^{n} \pi_i P_i$ and $\pi_1, \ldots, \pi_c$ are weights for probability distributions and $H(.)$ is the Shannon entropy.

In this study, JSD is computed for all pairs of classes, resulting in a pairwise dissimilarity matrix.

**Task-Similarity Distance:**

Noting that a reference distribution, which is the average of compared distributions is used in JSD, alternative reference distributions such as classification distributions (Gao & Chaudhari, 2020; Helm et al., 2020) are developed with the recent progress in transfer learning (Pan & Yang, 2009). For the purpose of using conditional distribuitons only, these innovative similarity measures are leveraged by introducing a reference conditional distribution and artificially constructing two classification distributions. TSD revolves around task similarity (Helm et al., 2020) as a metric employed to induce a hierarchy.

Specifically, in measuring the similarity between two conditional distributions $P$ and $Q$ using task similarity $TS$ defined by Helm et al. 2020, a third distribution $M$ and two classification distributions $P'$ and $Q'$ are defined as:

$$P' = p(X|Y \sim P)P + (1 - p(X|Y \sim P))M \quad [4]$$

$$Q' = p(X|Y \sim Q)Q + (1 - p(X|Y \sim Q))M \quad [5]$$

The similarity is then calculated as:

$$s(P, Q) = \frac{1}{2}\big(TS(P', Q') + TS(Q', P')\big) \quad [6]$$

In scenarios with large multi-class problems, $M$ is randomly sampled from the set of conditional distributions with $P$ and $Q$ excluded. Subsequently, multiple instances of $M_i$ are randomly chosen in practice, and the average similarity $s(P, Q)$ is computed to form a pairwise similarity matrix. Consequently, similarity emerges as a stochastic measure. It is worth noting that $s(P, Q)$ is asymmetrical and scaled. Therefore, when constructing the dissimilarity matrix, symmetry is achieved by averaging it with its transpose, and all values are normalized to fall within the range of 0 to 1. Ultimately, similarity is transformed into dissimilarity by subtracting non-diagonal values from 1.

**Classifier Based Distance:**

In essence, when a classifier struggles to differentiate between two classes, it suggests their similarity, while clear differentiation implies dissimilarity. The conventional method of assessing a classifier's ability to distinguish between two classes is by evaluating its performance. Therefore, examining the training data with a classifier yields dissimilarity measures between the classes in a supervised way.

One option is to analyze separability between all class pairs. This option can can become extremely time-consuming if the number of classes is high. Practitioners avoid this problem by obtaining a confusion matrix using faster approaches like single classifier. One option is to use the confusion matrix's rows as the corresponding class representatives (Bengio et al., 2010; Godbole, 2002). Another approach introduced by Silva-palacios et al. 2018 is to calculate the dissimilarity between two classes as the accuracy measured in the corresponding subset of the confusion matrix $M$.

$$M = \begin{bmatrix} m_{11} & \cdots & m_{1c} \\ \vdots & \ddots & \vdots \\ m_{c1} & \cdots & m_{cc} \end{bmatrix} \quad [7]$$

Using the confusion matrix elements, dissimilarity between two classes can be obtained as:

$$d(c_i, c_j) = \frac{m_{ii} + m_{jj}}{m_{ij} + m_{ji} + m_{ii} + m_{jj}}$$

[8]

**Time Series Datasets:**

The effectiveness of the proposed approach is evaluated using a subset of datasets sourced from the UCR archive (Dau et al., 2019). This archive comprises 128 univariate time series datasets. Specifically, this study focuses on multi-class cases with more than two classes. Moreover, datasets requiring over 2 minutes for training in a single fold with MINIROCKET are excluded to ensure efficiency. Following these criteria, a total of 46 datasets out of the original 128 are selected for evaluation. Detailed information regarding the selected datasets is provided in Table 1 in the Appendix section. These datasets are managed and analyzed using the sktime library (Löning et al., 2019), which offers specialized tools for handling time series data. They are utilized to evaluate the performance of the proposed hierarchical classification approach.

**Performance Evaluation and Statistical Analysis:**

The classification performance is assessed using the F1 score, a commonly employed metric for binary classification tasks. In the context of multi-class classification, the F1 macro score is adopted. This score computes the F1 score for each class and then calculates the average, assigning equal importance to each class. By employing both the F1 score and F1 macro score as evaluation metrics, the study comprehensively and robustly evaluates the classification performance of the selected classifiers.

Performance assessment is conducted using 5-fold cross-validation. The classification performance across all scenarios is visualized using a critical difference diagram (Ismail Fawaz et al., 2019). The null hypothesis assumes equal performance among all algorithms. To examine this hypothesis, the Friedman test (Friedman, 1940) is employed, followed by pairwise post-hoc analysis (Benavoli et al., 2016) using the Wilcoxon signed-rank test with Holm's alpha correction (Garcia and Herrera, 2008). A significance level of 5% is applied for alpha correction.

**RESULTS AND DISCUSSIONS**

The study investigated the performance of HC for time series data using various dissimilarity approaches, including JSD, TSD, and CBD. The effectiveness of these approaches was evaluated through experiments employing different classifier models: MINIROCKET, STSF, and SVM. MINIROCKET and STSF were selected as state-of-the-art time series classification algorithms, while SVM served as a standard machine learning algorithm. The configuration included 512 kernels for MINIROCKET (contrasting with its default size of around 10K) and 50 estimators for STSF, with a linear kernel chosen for SVM. The comparison results were visualized in Figure 2 using a critical difference diagram.

Based on the results, the primary finding indicates that HC exhibited significantly superior performance compared to FC when utilizing MINIROCKET with TSD. Conversely, when employing other classifiers such as STSF and SVM, FC demonstrated notably better performance across all configurations compared to HC. This study provides novel evidence highlighting the notable superiority of HC over FC for the first time.

Additionally, TSD consistently outperformed both CBD and JSD across nearly all scenarios, except for cases involving the STSF classifier where CBD outperformed TSD and JSD. The suboptimal performance of JSD can be attributed to its nature as a distribution-based dissimilarity measure that does not account for temporal relations between data instances.

The study's findings underscore the nuanced interplay between hierarchical and flat classification methodologies when applied to time series data. Notably, the superiority of HC over FC is evident when employing MINIROCKET with TSD. This observation marks a significant departure from previous understandings and emphasizes the effectiveness of HC in certain configurations.

Conversely, when utilizing alternative classifiers such as STSF and SVM, FC consistently outperforms HC across all configurations. This unexpected outcome challenges conventional assumptions about the comparative efficacy of hierarchical versus flat classification methods, particularly in the context of time series data analysis.

Further analysis reveals TSD as a consistently robust dissimilarity measure, outperforming both CBD and JSD in the majority of scenarios. However, an exception arises when employing the STSF classifier, where CBD displays superior performance to both TSD and JSD. This discrepancy highlights the nuanced nature of dissimilarity measures and underscores the importance of selecting an appropriate measure tailored to the specific characteristics of the dataset and classifier employed.

Overall, these findings contribute valuable insights into the performance dynamics of hierarchical and flat classification approaches in the context of time series data analysis. The study's results pave the way for future research aimed at refining classification methodologies and dissimilarity measures to optimize performance across diverse datasets and analytical scenarios.

**Limitations and Future Directions:**

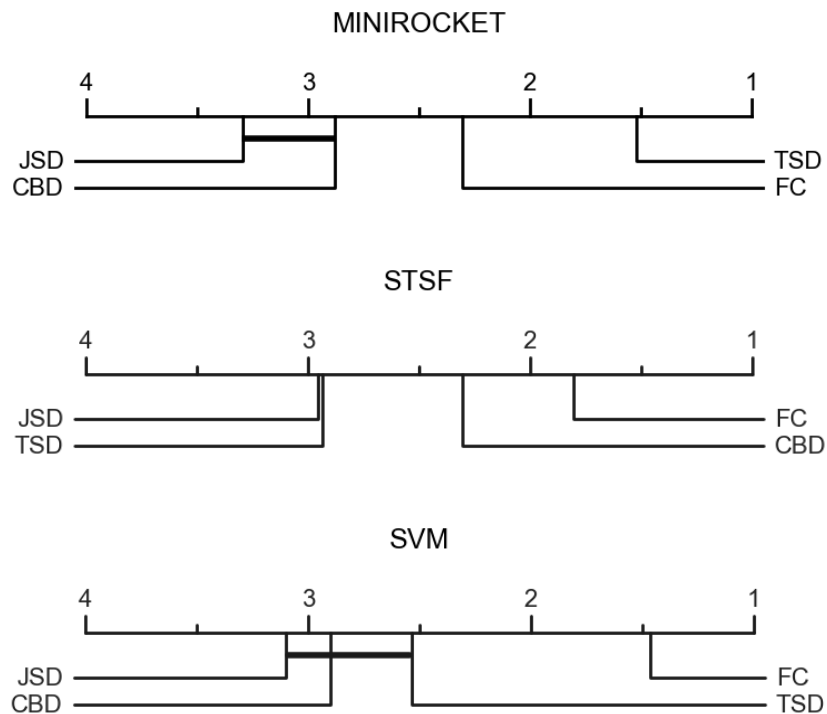

Figure 2. Critical difference diagram displays the ranking of HC performances when dissimilarity measure is varied in case of different classifier models.

*Complexity of Hierarchy Quality Assessment:* The study primarily relies on the assessment of hierarchy quality through the performance of HC. However, hierarchy quality is acknowledged as a complex and multifaceted concept. The current approach does not provide a straightforward, exhaustive measure of the quality of the generated hierarchy. Future studies could explore alternative unsupervised methods for assessing hierarchy quality, considering collective cluster qualities within the hierarchy to offer a more comprehensive evaluation.

*Limited Hierarchy Exploitation Approaches*: The hierarchy exploitation aspect in this study is constrained to a local classifier per node approach. While this choice allows for detailed examination, it limits the exploration of alternative exploitation strategies. Consideration of global classification or other local classification approaches, such as employing local classifiers per level or per node, could provide a more nuanced understanding of how different exploitation strategies impact overall HC performance.

*Pre-processing Considerations*: The study does not delve into the impact of pre-processing steps, such as dimension reduction, on the data samples before obtaining dissimilarity measures. Exploring the effects of various pre-processing techniques on the performance of dissimilarity measures and subsequent HC could contribute to a more comprehensive analysis.

*Alternative Dissimilarity Measures*: The comparison is focused on three dissimilarity measures (JSD, TSD, and CBD). While these were chosen deliberately, the study does not explore the potential benefits or drawbacks of other dissimilarity measures. Specifically, investigating class-conditional centroid approaches and their comparison with the selected dissimilarity measures could provide additional insights into the hierarchy generation process.

Addressing these limitations would contribute to a more comprehensive understanding of the nuances associated with dissimilarity measures, hierarchy quality assessment, and hierarchy exploitation in the context of HC for time series data.

**Conclusions:**

In conclusion, this study sheds light on the performance dynamics of hierarchical and flat classification methods in the realm of time series data analysis. The findings highlight a notable superiority of Hierarchical Classification (HC) over Flat Classification (FC) specifically when MINIROCKET is employed with Task Similarity Distance (TSD). This novel observation underscores the efficacy of HC in certain configurations, challenging conventional assumptions about classification methodologies.

Conversely, FC consistently outperforms HC across all configurations when utilizing alternative classifiers such as STSF and SVM. This unexpected outcome emphasizes the need for careful consideration and selection of classification methods based on the specific characteristics of the dataset and the underlying analytical goals.

Moreover, Task Similarity Distance (TSD) emerges as a robust dissimilarity measure, consistently outperforming both Classifier Based Distance (CBD) and Jensen-Shannon Distance (JSD) in the majority of scenarios. However, the superior performance of CBD over TSD and JSD in scenarios involving the STSF classifier highlights the nuanced nature of dissimilarity measures and the importance of their tailored selection.

Overall, these findings contribute valuable insights into the comparative effectiveness of hierarchical and flat classification methods in the context of time series data analysis. They underscore the need for further research aimed at refining classification methodologies and dissimilarity measures to optimize performance across diverse datasets and analytical scenarios.

**APPENDIX**

Table 1. Multi-class time series datasets from UCR archive.

|    | Data Type | Dataset | c | N | Length |
|----|-----------|---------|---|---|--------|
| 0  | Image     | Adiac   | 37 | 781 | 176 |
| 1  | Image     | ArrowHead | 3 | 211 | 251 |
| 2  | Spectro   | Beef    | 5 | 60  | 470 |
| 3  | Sensor    | Car     | 4 | 120 | 577 |
| 4  | Sensor    | ChlorineConcentration | 3 | 4307 | 166 |
| 5  | Motion    | CricketX | 12 | 780 | 300 |
| 6  | Motion    | CricketY | 12 | 780 | 300 |
| 7  | Motion    | CricketZ | 12 | 780 | 300 |
| 8  | Image     | DiatomSizeReduction | 4 | 322 | 345 |
| 9  | Image     | DistalPhalanxOutlineAgeGroup | 3 | 539 | 80 |
| 10 | Image     | DistalPhalanxTW | 6 | 539 | 80 |
| 11 | ECG       | ECG5000 | 5 | 5000 | 140 |
| 12 | Image     | FaceAll | 14 | 2250 | 131 |
| 13 | Image     | FaceFour | 4 | 112 | 350 |
| 14 | Image     | FacesUCR | 14 | 2250 | 131 |
| 15 | Image     | FiftyWords | 50 | 905 | 270 |
| 16 | Image     | Fish    | 7 | 350 | 463 |
| 17 | Motion    | Haptics | 5 | 463 | 1092 |
| 18 | Motion    | InlineSkate | 7 | 650 | 1882 |
| 19 | Sensor    | InsectWingbeatSound | 11 | 2200 | 256 |
| 20 | Device    | LargeKitchenAppliances | 3 | 750 | 720 |
| 21 | Sensor    | Lightning7 | 7 | 143 | 319 |
| 22 | Simulated | Mallat  | 8 | 2400 | 1024 |
| 23 | Spectro   | Meat    | 3 | 120 | 448 |
| 24 | Image     | MedicalImages | 10 | 1141 | 99 |
| 25 | Image     | MiddlePhalanxOutlineAgeGroup | 3 | 554 | 80 |
| 26 | Image     | MiddlePhalanxTW | 6 | 553 | 80 |
| 27 | Spectro   | OliveOil | 4 | 60 | 570 |
| 28 | Image     | OSULeaf | 6 | 442 | 427 |
| 29 | Image     | ProximalPhalanxOutlineAgeGroup | 3 | 605 | 80 |
| 30 | Image     | ProximalPhalanxTW | 6 | 605 | 80 |
| 31 | Device    | RefrigerationDevices | 3 | 750 | 720 |
| 32 | Device    | ScreenType | 3 | 750 | 720 |
| 33 | Device    | SmallKitchenAppliances | 3 | 750 | 720 |
| 34 | Image     | SwedishLeaf | 15 | 1125 | 128 |
| 35 | Image     | Symbols | 6 | 1020 | 398 |
| 36 | Simulated | SyntheticControl | 6 | 600 | 60 |
| 37 | Image     | WordSynonyms | 25 | 905 | 270 |
| 38 | Motion    | Worms   | 5 | 258 | 900 |
| 39 | Device    | ACSF1   | 10 | 200 | 1460 |
| 40 | Spectro   | EthanolLevel | 4 | 1004 | 1751 |
| 41 | Spectrum  | Rock    | 4 | 70 | 2844 |
| 42 | Spectrum  | SemgHandMovementCh2 | 6 | 900 | 1500 |
| 43 | Spectrum  | SemgHandSubjectCh2 | 5 | 900 | 1500 |
| 44 | Simulated | SmoothSubspace | 3 | 300 | 15 |
| 45 | Simulated | UMD     | 3 | 180 | 150 |

c: Number of flat classes, N: Number of samples